# Un résumeur à base de graphes, indépéndant de la langue

*Juan-Manuel Torres-Moreno*
*Laboratoire Informatique d'Avignon (LIA)*
*Université d'Avignon et des Pays de Vaucluse*
*Avignon, France*
*juan-manuel.torres@univ-avignon.fr*

*Javier Ramirez*
*Departamento de Sistemas / CBI*
*Universidad Autonoma Metropolitana-Azcapotzalco*
*Mexico DF, Mexique*

*Iria da Cunha*
*Laboratoire Informatique d'Avignon (LIA) UAPV*
*Institut Universitari de Lingüística Aplicada, UPF, Espagne*
*iria.dacunha@upf.edu*

***Abstract*—*In this paper we present REG, a graph-based approach for study a fundamental problem of Natural Language Processing (NLP): the automatic text summarization. The algorithm maps a document as a graph, then it computes the weight of their sentences. We have applied this approach to summarize documents in three languages.***

***Keywords: Automatic text summarization; Graphs algorithms; NLP; Cortex***

## 1. INTRODUCTION

Le résumé automatique de documents est une des méthodes de fouille de texte qui permet de compresser un document avec perte d'information, tout en conservant son informativité. Il s'agit d'une problématique importante du Traitement Automatique de Langues (TAL). Résumer consiste à condenser l'information la plus importante issue d'un ou plusieurs documents, afin d'en produire une version abrégée de son contenu [1]. Les gros titres des nouvelles, les bandes annonces et les synopsis sont quelques exemples de résumés utilisés couramment. De manière générale, les personnes sont des résumeurs extrêmement performants. Les approches par extraction de phrases combinant des algorithmes numériques et statistiques ont montré leur pertinence dans cette tâche. En se basant sur les études du comportement des résumeurs professionnels et notamment sur les travaux de [2,3], les chercheurs ont essayé d'imiter le processus cognitif de création d'un résumé. Les premiers travaux portant sur le résumé automatique de textes datent de la fin des années 50 [4]. Luhn décrit une technique simple, spécifique aux articles scientifiques qui utilise la distribution des fréquences de mots dans le document pour pondérer les phrases. Luhn était déjà motivé par la problématique de surcharge d'information. Il décrit quelques uns des avantages qu'ont les résumés produits de manière automatique par rapport aux résumés manuels : coût de production très réduit, non assujetti aux problèmes de subjectivité et de variabilité observés sur les résumeurs professionnels. L'idée de Luhn d'utiliser des techniques statistiques pour la production automatique de résumés a eu un impact considérable, la grande majorité des systèmes d'aujourd'hui étant basés sur ces mêmes idées.

Par la suite, [5] a étendu les travaux de Luhn en tenant compte de la position des phrases, de la présence des mots provenant de la structure du document (par exemple les titres, sous-titres, etc.) et de la présence de mots indices (*significant*, *impossible*, *hardly*, etc.). Les recherches menées par [6] au sein du *Chemical Abstracts Service* (CAS) dans la production de résumés à partir d'articles scientifiques de Chimie ont permis de valider la viabilité des approches d'extraction automatique de phrases. Un nettoyage des phrases reposant sur des opérations d'élimination fut pour la première fois introduit. Les phrases commençant par exemple par « *in* » (par exemple « *in conclusion* ») ou finissant par « *that* » seront éliminées du résumé. Afin que les résumés satisfassent les standards imposés par le CAS, une normalisation du vocabulaire est effectuée. Elle inclut le remplacement des mots/phrases par leur abréviation, une standardisation des variantes orthographiques. Ces travaux ont posé les bases du résumé automatique de textes. Une méthodologie de production des résumés émerge de leur analyse : i) Prétraitement, ii) Identification des phrases saillantes dans le document source, iii) Construction du résumé par concaténation des phrases extraites, iv) Traitement surfacique des phrases.

Ce travail porte sur la conception d'un système de résumé automatique générique. Ce système utilise des méthodes de graphes afin de repérer dans le document les phrases les plus importantes. Notre volonté de n'utiliser que des traitements statistiques est motivée par le fait que le système doit être le plus indépendant possible de la langue. La méthode proposée repose sur un prétraitement des documents et sur une fonction de pondération des phrases par optimisation d'un graphe.

## 2. ALGORITHMES DE RÉSUMÉ À BASE GRAPHES

[7,8] considèrent le résumé par extraction comme une identification des segments les plus *prestigieux* dans un graphe. Les algorithmes de classement basés sur les graphes tel que [9] ont été utilisés avec succès dans les réseaux sociaux, l'analyse du nombre de citations ou l'étude de la structure du Web. Ces algorithmes peuvent être vus comme les éléments clés du paradigme amorcé dans le domaine de la recherche sur Internet, à savoir le classement des pages Web par l'analyse de leurs positions dans le réseau et non pas de leurs contenus. En d'autres termes, ces algorithmes permettent de décider de l'importance du sommet d'un graphe en se basant non pas sur l'analyse locale du sommet lui même, mais sur l'information globale issue de l'analyse récursive du graphe complet. Appliqué au résumé automatique cela signifie que le document est représenté par un graphe d'unités textuelles (phrases) liées entre elles par des relations issues de calculs de similarité. Les phrases sont ensuite sélectionnées selon des critères de centralité ou de prestige dans le graphe puis assemblées pour produire des extraits. Les résultats reportés montrent que les performances des approches à base de graphe sont au niveau des meilleurs systèmes actuels [10] mais ne portent que sur des documents en anglais et en portugais. Il est important de noter que les méthodes de

classement sont entièrement dépendantes de la bonne construction du graphe sensé représenter le document. Puisque ce graphe est généré à partir de mesures de similarités inter-phrases, l'impact que peut avoir le choix de la méthode de calcul est à considérer. Dans leurs travaux, [7,8] utilisent le modèle en sac-de-mots pour représenter chaque phrase comme un vecteur à *N* dimensions, (*N*=Nb total de mots différents) et chaque composante du vecteur un poids *tf idf*. Les valeurs de similarité entre phrases sont ensuite obtenues par un calcul du cosinus entre leurs représentations vectorielles. Le point faible de cette mesure, et plus généralement de toutes les mesures utilisant les mots comme unités, est qu'elles sont tributaires du vocabulaire. Dans une optique d'indépendance de la langue, les prétraitements qui sont appliqués aux segments se doivent d'être minimaux. C'est malheureusement dans cette configuration que les performances de la mesure cosinus chutent car elle ne permet en aucun cas de mettre en relation des mots qui morphologiquement peuvent être très proches. Une solution combinant les mesures de similarité et celles basées sur les caractères. [11] proposent une mesure dérivée d'un calcul de similarité entre chaînes de caractères originellement employé pour la détection d'entités redondantes (*Record Linkage*). Cette mesure permet de créer des relations entre deux segments qui même s'ils ne partagent aucun mot, en contiennent des morphologiquement proches. Une seconde question est donc de savoir si la construction du graphe du document à partir de mesures mixtes (mots et caractères) permet d'améliorer l'extraction de segments. [12] ont montré que cela est possible. Nous voulions cependant une solution avec un algorithme à base de graphes encore plus simple. Nous posons le problème du résumé automatique de texte par extraction comme un problème d'optimisation. Ainsi, un texte est représenté comme un graphe non dirigé qui peut être assimilé comme un problème de coloration ou à une des variantes de celui du voyageur du commerce. Le problème ainsi posé est de l'ordre de *P!*, étant *P* le nombre de phrases d'un document. Cela fait de cette tâche un problème *NP*-complet. Nous nous sommes tournés vers les approches gloutonnes. Nous avons développé un algorithme optimal de visite des *m* sommets, *m* fixé par l'utilisateur. L'algorithme REG (REsumeur à base de Graphes) réalise une l'extraction des phrases les plus pertinentes d'un résumé par extraction.

## 3. REG : UN ALGORITHME RESUMEUR À BASE DE GRAPHES

REG consiste en deux grandes phases : d'abord une représentation adéquate des documents, puis une pondération des phrases. La première est réalisée au moyen d'une représentation vectorielle qui est assez indépendante de la langue. La deuxième par un algorithme d'optimisation glouton. La génération du résumé est effectuée par concaténation des phrases pertinentes, pondérées dans l'étape d'optimisation.

### *3.1 Pré-traitement et représentation vectorielle*

Les documents sont pré-traités avec des algorithmes classiques de filtrage de mots fonctionnels (avec un d'anti-dictionnaire), de normalisation et de lemmatisation [1,13] afin de réduire la dimensionnalité. Une représentation en sac de mots produit une matrice $S_{[P \times N]}$ de fréquences/absences composée de $\mu=1,\dots,P$ phrases (lignes) ; $\sigma_\mu = s_{\mu,1},\dots,s_{\mu,i},\dots,s_{\mu,N}$ et un vocabulaire de $i=1,\dots,N$ termes (colonnes).

(1) $$S = \begin{pmatrix} s_{1,1} & s_{1,2} & \cdots & s_{1,N} \\ s_{2,1} & s_{2,2} & \cdots & s_{2,N} \\ \vdots & \vdots & \ddots & \vdots \\ s_{P,1} & s_{P,2} & \cdots & s_{P,N} \end{pmatrix} \; ; \quad s_{\mu,i} = \begin{cases} TF_i & \text{si le terme } i \text{ existe} \\ 0 & \text{autrement} \end{cases}$$

La présence du mot *i* est représentée par sa fréquence TF_*i* (son absence par 0 respectivement), et une phrase $\sigma_{\mu}$ est donc un vecteur de *N* occurrences. *S* est une matrice entière car ses éléments prennent des valeurs fréquentielles absolues.

## *3.2 Une solution gloutonne*

A partir du modèle vectoriel de représentation de documents, nous proposons de créer un graphe *G = (S, A)* où les sommets *S* représentent les phrases et *A* l'ensemble d'arêtes. Une arête entre deux sommets est créée si les phrases correspondantes possèdent au moins un mot en commun. On construit une matrice d'adjacence à partir de la matrice $S_{\text{[phrases x mots]}}$ comme suit : Si l'élément *S{i,k}*= 1 de la matrice *S* (dans la phrase *i* le mot *k* est présent), on vérifie dans la colonne *k* et quand un élément *S{j,k}=1* on met 1 dans la case *a{i,j}* de la matrice d'adjacence *A*, qui veut que les phrases *i* et *j* partagent le mot *k*. Pour afficher les phrases les plus lourdes nous avons trouvé qu'il fallait chercher une variante du problème de l'arbre de poids maximum, où les poids sont sur les sommets, pas sur les arêtes. Nous avons ainsi construit un algorithme inspiré de l'algorithme de Kruskal [14]. L'algorithme proposé fonctionne de la façon suivante : i) générer la matrice d'adjacence *A* qui aura autant des lignes et des colonnes que des phrases considérées, c'est à dire *P* ; ii) calculer le poids des sommets : la somme d'arêtes entrantes du sommet ;iii) calculer le degré de chaque sommet : le nombre de mots partagé avec les autres phrases ; iv) demander le pourcentage *k* des phrases qui aura le résumé et le déterminer. La matrice d'adjacence $A_{\text{[P x P]}}$ sera générée à partir de la représentation vectorielle (voir l'équation (1). Le calcul est comme suit : parcourir la ligne *i*, *i*=1,…,*P*, et chaque élément *a{i,j}* égal à 1 descendre par la colonne *j* pour identifier d'autres phrases qui partagent ce mot :

$$
S^{\mathrm{Mars}}_{[P=11\times N=16]} = \begin{pmatrix}
0&0&0&0&0&1&0&0&0&0&0&0&0&1&0&0\\
1&1&0&0&0&0&1&1&0&0&0&0&0&0&1&0\\
0&0&0&1&0&0&0&0&0&0&1&0&0&0&0&0\\
0&0&1&0&0&0&1&1&1&1&0&0&0&0&1&1\\
0&0&0&0&0&1&0&0&1&0&0&0&0&0&0&1\\
1&0&0&0&0&0&0&0&0&0&0&0&0&0&0&0\\
0&1&0&0&0&0&0&0&0&0&0&0&0&0&0&0\\
0&0&0&0&0&0&1&1&0&1&0&1&1&1&0&0\\
0&0&1&1&0&0&1&0&0&0&0&0&0&0&0&1\\
0&0&0&0&1&0&0&0&0&0&0&1&1&0&1&0\\
1&0&0&0&0&0&0&0&0&0&0&0&0&0&0&0
\end{pmatrix}
\quad \leftarrow \text{phrase } i \qquad (\text{mot } j \text{ : colonne } 8 \downarrow)
$$

Les poids des 11 sommets (représentant les phrases) du graphe correspondant sont : a0 = 2, a1 = 5, a2 = 2, a3 = 7, a4 = 3, a5 = 1, a6 = 1, a7 = 7, a8 = 4, a9 = 4, a10 = 1. Cela donne lieu à une matrice d'adjacence. Nous montrons le fonctionnement de l'algorithme sur le graphe correspondant (voir la figure 1).

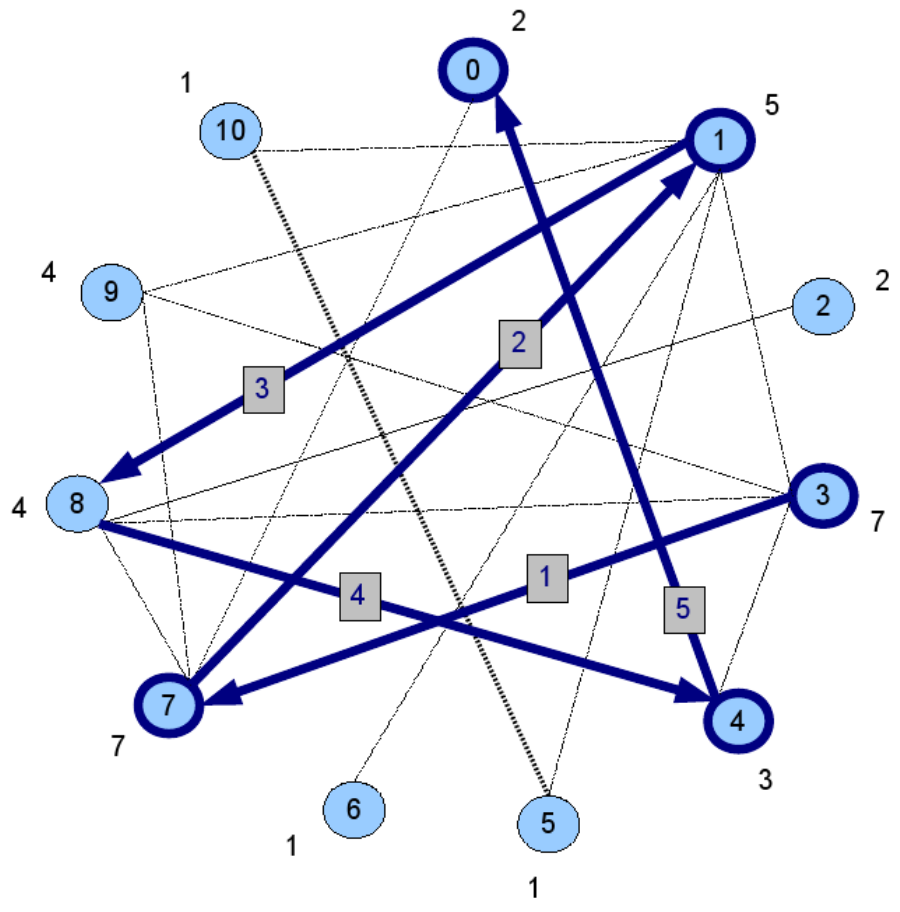

Figure 1 Graphe correspondant au texte « Mars ».

## 4 EXPÉRIENCES SUR LE RÉSUMÉ AUTOMATIQUE

Sous l'hypothèse que le poids d'une phrase μ indique son importance dans le document, nous avons appliqué l'algorithme REG au résumé par extraction de phrases [1,15]. Notre méthode est orientée, pour le moment, à la génération de résumés génériques mono-document. Cependant, nous pensons qu'une modification simple de l'approche (voir conclusion) pourrait nous permettre d'obtenir des résumés multi-document guidés par une requête ou un sujet défini par l'utilisateur (ce qui correspond au protocole des conférences DUC/TAC (*Document Understandig Conferences* http://www-nlpir.nist.gov/projects/duc/index.html). L'algorithme REG de résumé automatique comprend trois modules. Le premier réalise la transformation vectorielle du texte avec des processus de filtrage, de lemmatisation/*stemming* et de normalisation. Le second module applique l'algorithme glouton et réalise le calcul de la matrice d'adjacence. Nous obtenons la pondération de la phrase $\nu$ directement de l'algorithme. Ainsi, les phrases pertinentes seront sélectionnées comme ayant la plus grande pondération. Finalement, le troisième module génère les résumés par affichage et concaténation des phrases pertinentes. Le premier et le dernier module reposent sur le système Cortex [16,19]. Nous avons évalué les résumés produits par notre système avec le logiciel ROUGE [17], qui mesure la similarité, suivant plusieurs stratégies, entre un résumé candidat (produit automatiquement) et des résumés de référence (créés par des humains).

Nous avons réalisé une batterie de tests différents sur un corpus de documents très hétéroclite (732 phrases, 18 270 mots). Des évaluation de textes en français (domaine ouvert, textes composites et littéraire) ; textes encyclopédiques en anglais ; et textes en espagnol d'un domaine de spécialité. Pour l'évaluation des tests en français (récupérables sur le site http://www.lia.univ-avignon.fr) nous avons choisi le corpus suivant : « Mars », « Puces » et la lettre d'Emile Zola, « J'accuse » http://fr.wikipedia.org/wiki/J'accuse...!. Deux textes de la Wikipédia en anglais ont été analysés: « Lewinksky » http://en.wikipedia.org/wiki/Monica_Lewinsky, et « Québec » http://en.wikipedia.org/wiki/ Quebec_sovereignty_movement. Enfin, en espagnol nous avons utilisé des textes de la revue *Medicina Clinica* http://www.elsevier.es/revistas/ ctl_servlet?f=7032&revistaid=2. Pour cette tâche, un corpus composé de 8 textes (~ 400 phrases et 11 000 mots) a été sélectionné. Nous avons évalué les résumés produits par notre système avec ROUGE. Dans le cas des corpus français et anglais, les résumés de référence ont été produits par plusieurs juges de niveau d'études universitaire. Pour le corpus en espagnol, nous avons utilisé les résumés produits par les auteurs comme résumé de référence. Dans la table 1 nous présentons le détail de mesures Rouge-2 et SU4 pour le texte Mars. Dans cette table on constate que les trois premières places sont *ex-equo* par REG, Cortex et Enertex [18].

| Rouge | REG | Cortex | Enertex | OTS | Copernic | Word | Random | Leadbase | Pertinence | Swseum |
|---|---|---|---|---|---|---|---|---|---|---|
| -2 | **0.8198** | **0.8198** | **0.8198** | 0.5554 | 0.6620 | 0.1609 | 0.3923 | 0.3262 | 0.5117 | 0.0319 |
| SU4 | **0.8128** | **0.8128** | **0.8128** | 0.5716 | 0.6729 | 0.1627 | 0.3999 | 0.3271 | 0.5371 | 0.0560 |

Table 1, Mesures ROUGE pour le texte « Mars »

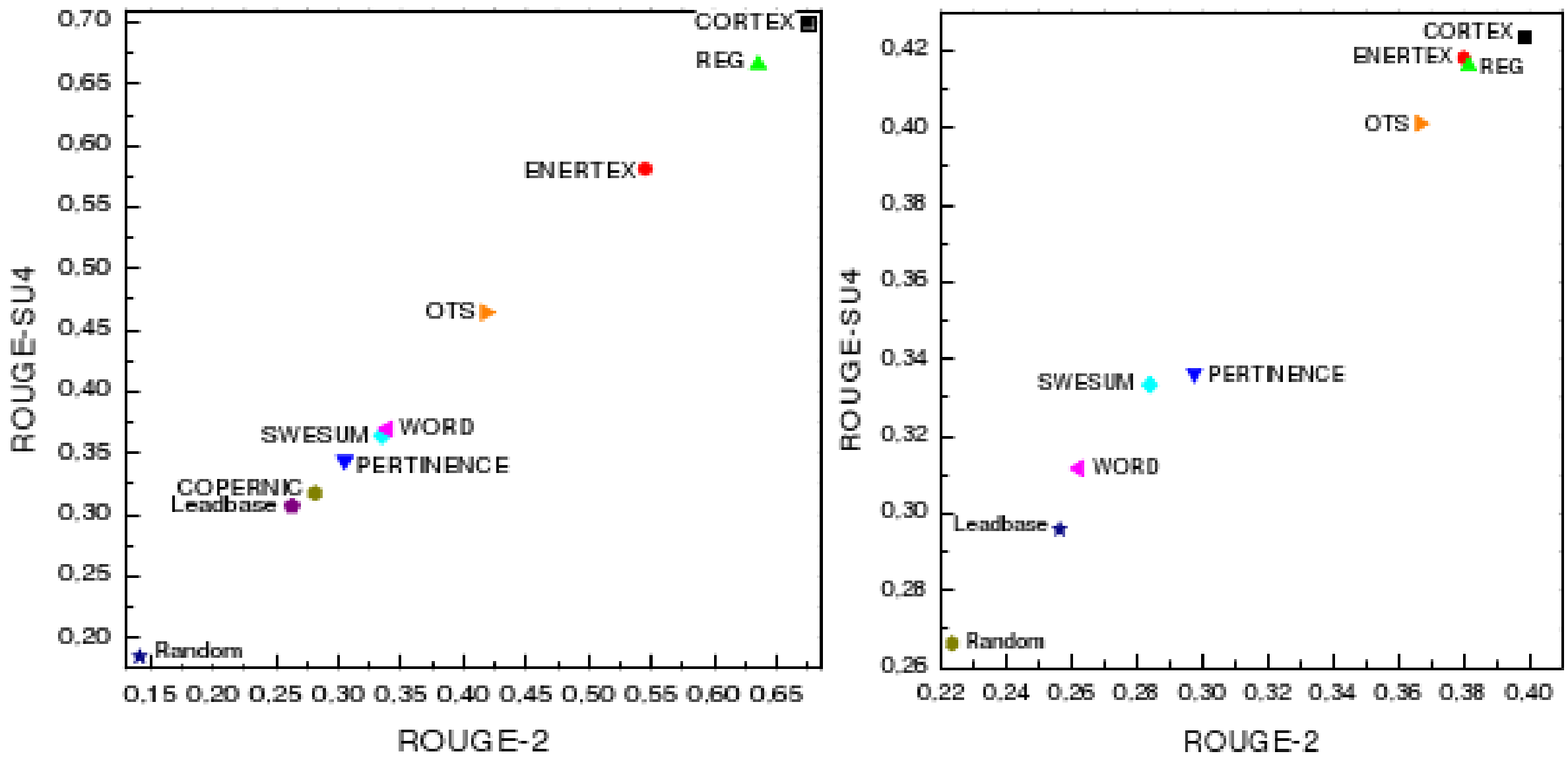

Figure 2. Mesures ROUGE pour « J'accuse » à gauche et sur le corpus de « *Medicina clinica* ».

## 5 CONCLUSION

Dans cet article nous avons introduit un algorithme glouton basé sur des approches de graphes. Cela nous a permis de développer un nouvel algorithme de résumé automatique. Des tests effectués ont montré que notre algorithme est efficace pour la recherche de segments pertinents. On obtient des résumés équilibrés où la plupart des thèmes sont abordés dans le condensé final. Les avantages supplémentaires consistent à ce que les résumés sont obtenus de façon indépendante de la taille du texte, des sujets abordés, d'une certaine quantité de bruit et de la langue (sauf pour la partie prétraitement).

Les résultats ici présentés sont très encourageants. Nous réservons aussi une expérience d'évaluation sur des résumés tronqués à un nombre fixe de mots. Ceci lisserait le biais de segmentation par phrase induit par les systèmes TAL selon des critères arbitraires. Nous pensons que l'algorithme glouton REG pourrait être incorporé au système Cortex, où il jouerait le rôle d'une des métriques pilotées par un algorithme de décision. Ceci permettrait d'obtenir des résumés à l'aide d'une requête de l'utilisateur ou des résumés multi-documents. Une autre voie intéressante est d'introduire un vecteur des termes d'un texte décrivant une thématique (topique) qui sera introduit dans le graphe du document. Ainsi, les phrases du document pourraient, ou non, s'aligner selon leur degré de pertinence par rapport à la thématique. Ceci permettrait de générer des résumés personnalisés, telles que définis dans les tâches TAC/DUC. L'approche de graphes orientés sera aussi considérée à l'avenir pour créer une espèce de chaîne « conceptuelle » entre les phrases. D'autres applications envisagées de cet algorithme concernent l'indépendance de la langue. Ainsi, nous nous proposons d'étudier des langues très éloignées des langues européennes dans le plan syntaxique et grammaticale, telles que le Somali par exemple. Une collaboration avec le Centre de Recherche de Djibouti est en cours.

## ACKNOWLEDGMENT

Ce projet a été financé partiellement par le projet RPM2 ANR (France).